\documentclass[]{spie}  %>>> use for US letter paper
\usepackage{amsmath,amsfonts,amssymb}
\usepackage{graphicx}
\usepackage[colorlinks=true, allcolors=blue]{hyperref}
\usepackage{placeins} 
\title{End-to-end reconstruction of OCT optical properties and speckle-reduced structural intensity via physics-based learning}

\author[a]{Jinglun Yu}
\author[a]{Yaning Wang}
\author[a,b]{Wenhan Guo}
\author[a]{Yuan Gao}
\author[a,b]{Yu Sun}
\author[a]{Jin U. Kang}
\affil[a]{Department of Electrical and Computer Engineering, Johns Hopkins University, Baltimore, USA}
\affil[b]{Data Science and AI Institute, Johns Hopkins University, Baltimore, USA
}

\authorinfo{Jinglun Yu: E-mail: jyu146@jhu.edu}

\begin{document} 
\maketitle

\begin{abstract}
Inverse scattering in optical coherence tomography (OCT) seeks to recover both structural images and intrinsic tissue optical properties, including refractive index, scattering coefficient, and anisotropy. This inverse problem is challenging due to attenuation, speckle noise, and strong coupling among parameters. We propose a regularized end-to-end deep learning framework that jointly reconstructs optical parameter maps and speckle-reduced OCT structural intensity for layer visualization. Trained with Monte Carlo–simulated ground truth, our network incorporates a physics-based OCT forward model that generates predicted signals from the estimated parameters, providing physics-consistent supervision for parameter recovery and artifact suppression. Experiments on the synthetic corneal OCT dataset demonstrate robust optical map recovery under noise, improved resolution, and enhanced structural fidelity. This approach enables quantitative multi-parameter tissue characterization and highlights the benefit of combining physics-informed modeling with deep learning for computational OCT.
\end{abstract}

% Include a list of keywords after the abstract 

\section{INTRODUCTION}
\label{sec:intro}  % \label{} allows reference to this section

Optical coherence tomography (OCT) acquires high-resolution cross-sectional images by measuring backscattered interferometric signals\cite{wang2024automatic,Wang2024SubretinalOCT}. It has become a widely used modality for microstructure assessment, disease diagnosis, and biomedical imaging\cite{wang2024reimaginingpartialthicknesskeratoplasty,Yu2025TopologyDALK,Yi2025KalmanDALK, xu2023neural}. However, the contrast in OCT intensity arises from only a narrow range of variations in intrinsic refractive index (typically 1.3–1.5)\cite{wang2022depth}, and backscattered signals are significantly affected by attenuation and speckle noise during light propagation. As a result, using only the OCT intensity signal often struggles to recover detailed structural information, especially in complex tissues containing fuzzy boundaries and heterogeneous compositions\cite{wang2021optimized}.

To obtain the more accurate tissue anatomy, there has been increasing interest in recovering intrinsic optical properties from OCT intensity by solving the associated inverse problem. These parameters enable a more meaningful physical characterization of the tissue including refractive index, scattering coefficient, and anisotropy\cite{wang2024optical}. However, the inverse problem is severely ill-posed: multiple unknown parameters are strongly coupled, OCT signal amplitude decays with depth, noise propagates through the OCT signal formation process, and reconstructions require suitable priors for stability\cite{wang2023depth}. Traditional approaches typically rely on Beer–Lambert attenuation modeling, exponential fitting, or iterative regularization\cite{chang2009attenuation,vermeer2013depth,cannon2021layer}. While effective under restricted assumptions, these methods are noise-sensitive, often require segmentation or depth discretization, and tend to oversmooth or degrade spatial continuity\cite{li2020robust}.

Deep learning has recently been explored for OCT signal reconstruction and optical parameter characterization\cite{wang2025superresolutionopticalcoherencetomography}. However, supervised models require extensive datasets with ground-truth parameters, which are difficult to obtain experimentally, limiting generalization to diverse tissues or imaging conditions\cite{Zuo2024FishEyeUNet}.

To address these challenges and enable joint recovery of multi-parameter tissue maps, we propose a regularized end-to-end deep learning framework with a multi-channel encoder–decoder backbone, in which a physics-based OCT forward model is embedded as a differentiable constraint. From predicted refractive index, scattering coefficient, and anisotropy, the network computes theoretical OCT signals and enforces consistency with measured intensities, enabling simultaneous learning of tissue scattering behavior and speckle-reduced structural intensity maps. In a single pass, the framework provides noise-robust optical maps together with improved structural visibility for label-free layer visualization. Validation on Monte Carlo–simulated corneal datasets $(N > 500)$ demonstrates robust recovery of structural information from OCT intensity under noise.

\section{Methods}

\begin{figure} [ht]
   \begin{center}
   \begin{tabular}{c} %% tabular useful for creating an array of images 
   \includegraphics[height=10cm]{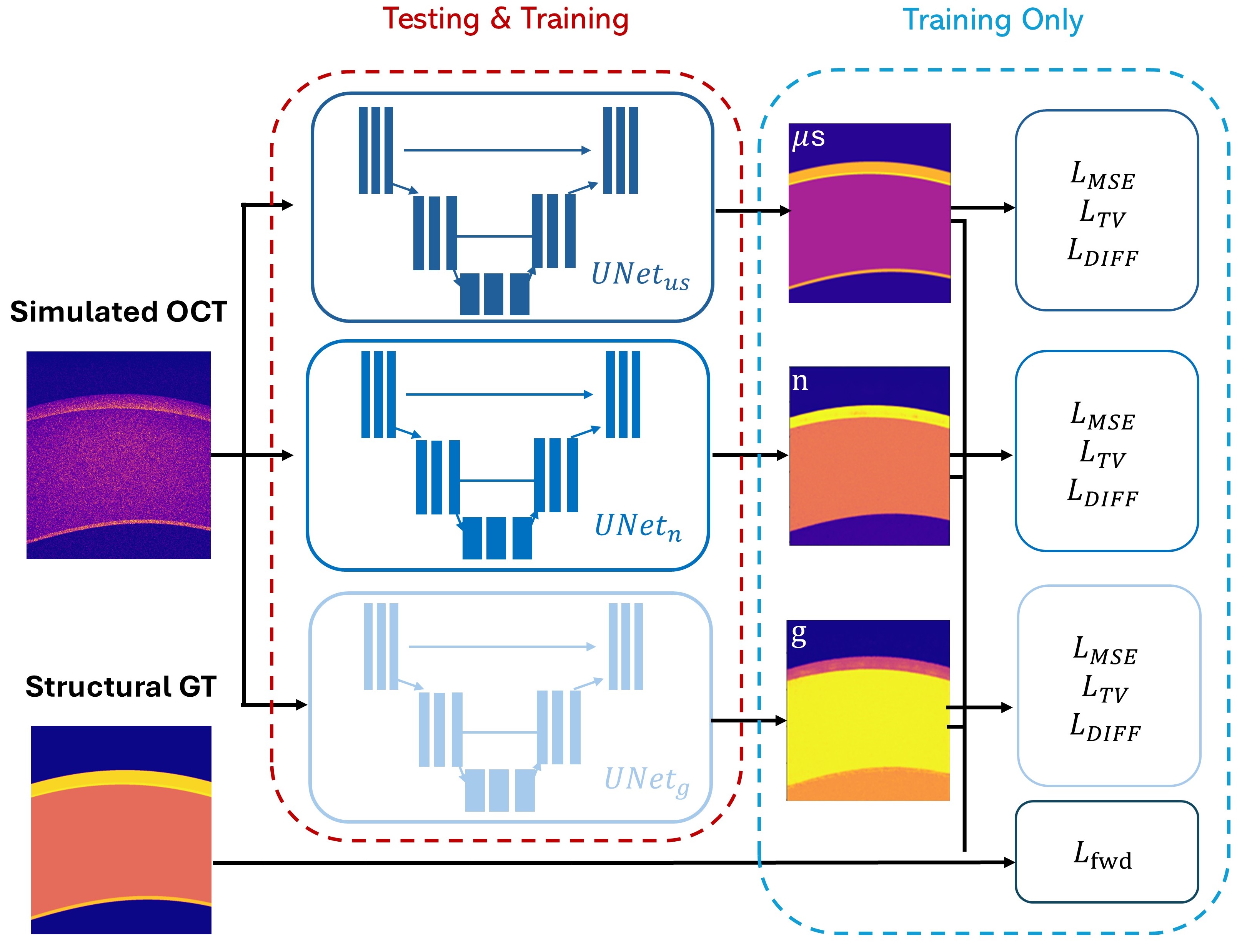}
   \end{tabular}
   \end{center}
   \caption[example] 
%>>>> use \label inside caption to get Fig. number with \ref{}
   { \label{fig:method} 
End-to-end physics-regularized framework for OCT inverse scattering. Red box: U-Net predictors used for both training and testing; blue box: loss modules applied only during training.
}
   \end{figure}

We propose an end-to-end OCT inverse reconstruction framework that performs single-pass inference to simultaneously recover intrinsic tissue optical properties and speckle-reduced OCT structural intensity, as shown in figure~\ref{fig:method}. The framework consists of two major components: (1) a U-Net-based multi-channel mapping network for optical properties reconstruction, and (2) a differentiable physics-based OCT forward model used as a consistency constraint during training.
\subsection{Network Architecture}
The network input is the raw OCT B-mode image $\mathbf{I}_{\mathrm{raw}}$ (Monte Carlo simulated $1024 \times 1024$). 
The dataset contains 500 corneal images synthesized based on prior geometric models of human corneal anatomy\cite{gerakis2008monte}. 
The network outputs three optical property maps---refractive index $n(x,y)$, scattering coefficient $\mu_s(x,y)$, and anisotropy $g(x,y)$---together with the speckle-reduced predicted OCT structural intensity $\hat{\mathbf{I}}(x,y)$.

The backbone adopts a U-Net-style encoder--decoder structure with four downsampling--upsampling stages and skip connections that preserve high-frequency spatial detailS. 
Convolution kernels of size $3 \times 3$ and BatchNorm layers are used in each stage; these hyperparameters were empirically tuned to achieve a balance between receptive-field size, computational cost, and training stability without oversmoothing. 
Each optical property is predicted through an independent U-Net branch rather than a shared backbone, ensuring that the optimization trajectories remain fully decoupled across $n$, $\mu_s$, and $g$. 
Although several loss terms (e.g., forward consistency) are applied simultaneously to all branches, network weights and gradients are not shared, preventing implicit interference between branch-specific representations.

\subsection{Physics-based Forward Model Constraint}
To establish a physical relationship between the estimated optical properties and the measured OCT intensity, we embed a differentiable OCT forward model into the training pipeline. The forward model is inspired by the Extended Huygens--Fresnel (EHF) theory for multiply scattered OCT signals, in which the mean squared OCT amplitude $A(z)$(equivalent to the OCT intensity $\hat{I}_{\mathrm{OCT}}$) along depth $z$ is composed of three contributions: (i) single scattering, (ii) multiple forward scattering, and (iii) the coherent cross-term between them\cite{gong2020parametric}.

Under the paraxial approximation and a highly forward-scattering regime ($g > 0.7$), the EHF formulation defines the local beam waist in the absence of forward scattering, denoted as $w_H(z)$, and its broadened counterpart $w_S(z)$ when multiple forward scattering is present. These quantities depend on system beam parameters, including the beam waist radius $w_0$, the Rayleigh length $z_R$, and the focal depth $z_f$, as well as on the anisotropy-dependent scattering angle
$
\theta_{\mathrm{RMS}} \approx \sqrt{2(1-g)} ,
$
which jointly result in depth-dependent beam propagation, reduced lateral resolution, and speckle blurring.

A differentiable approximation of the EHF forward model is implemented as

\begin{equation}
\left\langle A^{2}(z) \right\rangle 
\propto \frac{1}{w_H^{2}(z)}
\left\{
{
\exp\!\left(-2\mu_s z\right)
+ 4 \exp\!\left(-\mu_s z\right)\frac{\left[1 - \exp\!\left(-\mu_s z\right)\right]
}{1 +\frac{w_S^{2}(z)}{w_H^{2}(z)} }
}
+
\frac{
\left[1 - \exp\!\left(-\mu_s z\right)\right]^2
w_H^{2}(z)}{
w_S^{2}(z)
}
\right\},
\label{eq:ehf_forward}
\end{equation}
where,
\begin{equation}
\begin{aligned}
w_H^{2}(z) &= w_0^{2} \left[ \left( \frac{z - z_f}{2 n z_R} \right)^{2} + 1 \right], 
w_S^{2}(z) &= w_H^{2}(z) + \frac{1}{3} \left( \mu_s z \right) \theta_{\mathrm{RMS}}^{2} \left( \frac{z}{n} \right)^{2}.
\end{aligned}
\end{equation}
This formulation enables the network to simulate theoretical OCT intensities from the predicted optical properties $(n, \mu_s, g)$ and enforces physics-guided consistency with the measured OCT signals during training.

\subsection{Loss Function}
The total loss consists of four terms: 
\subsubsection{Parameter reconstruction loss}

The mean squared error (MSE) between the predicted optical properties $(n, \mu_s, g)$ and the Monte Carlo ground truth $(n^{*}, \mu_s^{*}, g^{*})$ :
\begin{equation}
\mathcal{L}_{\mathrm{MSE}}
=
\frac{1}{|\Omega|}
\sum_{(x,y)\in\Omega}
\left(
\left\| n(x,y) - n^{*}(x,y) \right\|_2^2
+
\left\| \mu_s(x,y) - \mu_s^{*}(x,y) \right\|_2^2
+
\left\| g(x,y) - g^{*}(x,y) \right\|_2^2
\right),
\label{eq:param_loss}
\end{equation}
where $\Omega$ denotes the spatial domain of the OCT Bscan image.

\subsubsection{Forward consistency loss}
Constraints forward-model output to match the measured OCT signal:
\begin{equation}
\mathcal{L}_{\mathrm{fwd}}
=
\frac{1}{|\Omega|}
\sum_{(x,y)\in\Omega}
\left\|
\hat{I}_{\mathrm{OCT}}(x,y)
-
I_{\mathrm{raw}}(x,y)
\right\|_2^2 ,
\label{eq:fwd_loss}
\end{equation}
where $\hat{I}_{\mathrm{OCT}} = \mathcal{F}_{\mathrm{EHF}}(n,\mu_s,g)$ denotes the OCT intensity generated by the physics-based forward model.
\subsubsection{Spatial regularization}
Total variation (TV) regularization is applied to suppress speckle noise in the raw OCT intensity $I_{\mathrm{raw}}$ and to encourage spatial smoothness and continuity in the predicted parameter maps:
\begin{equation}
\mathcal{L}_{\mathrm{TV}}
=
\sum_{p \in \{n,\mu_s,g\}}
\sum_{(x,y)\in\Omega}
\left(
\left|\nabla_x p(x,y)\right|
+
\left|\nabla_y p(x,y)\right|
\right),
\label{eq:tv_loss}
\end{equation}
where $\nabla_x$ and $\nabla_y$ denote spatial finite differences along the horizontal and vertical directions, respectively.
\subsubsection{Diffusion-based score regularization}
To further stabilize the inverse reconstruction, diffusion-model-based score regularization is incorporated as a learned statistical prior. The predicted parameter maps are first normalized and downsampled to $256\times256$, then perturbed by Gaussian noise following a predefined diffusion noise schedule. The resulting noisy samples are compared with score targets produced by pretrained EDM denoisers\cite{Karras2022EDMDesignSpace, guo2025psi3d}:
\begin{equation}
\mathcal{L}_{\mathrm{DIFF}}
=
\sum_{p \in \{n,\mu_s,g\}}
\omega_p
\,
\mathbb{E}_{\mathbf{x}_t}
\left[
\left\|
s_{\theta_p}(\mathbf{x}_t, t)
-
s^{*}(\mathbf{x}_t, t)
\right\|_2^2
\right],
\label{eq:diff_loss}
\end{equation}
where $s_{\theta_p}$ denotes the score function predicted by the pretrained EDM denoiser for parameter $p$, and $s^{*}$ is the analytical score target derived from the diffusion process.And $\omega_n=1$, $\omega_{\mu_s}=1$, $\omega_g=0.3$ are weighting coefficients that balance the contributions of each branch to stabilize gradient magnitudes during training. Because the EDM priors were trained on over 20,000 Monte Carlo scattering simulations, this regularization introduces a reliable statistical prior, improving robustness and convergence stability in the inverse mapping.
\subsubsection{Total loss}

The overall training objective is given as:
\begin{equation}
\mathcal{L}_{\mathrm{total}}
=
\lambda_1 \mathcal{L}_{\mathrm{MSE}}
+
\lambda_2 \mathcal{L}_{\mathrm{fwd}}
+
\lambda_3 \mathcal{L}_{\mathrm{TV}}
+
\lambda_4 \mathcal{L}_{\mathrm{DIFF}},
\label{eq:total_loss}
\end{equation}
where $\lambda_1$, $\lambda_2$, $\lambda_3$, and $\lambda_4$ are balancing weights.

\subsection{Training and Inference}
Training is performed using patch-based mini-batches on an NVIDIA A100 GPU, with learning rate $2  \times 10^{- 5}$ and Adam optimizer. Training is stopped when validation accuracy degrades. During inference, only a single forward pass is required to simultaneously produce speckle-reduced structural intensity and all three optical parameter maps.

\section{Results}
\label{sec:sections}

\subsection{Qualitative Comparisons}

\begin{figure}[ht]
    \centering
    \includegraphics[width=1\linewidth]{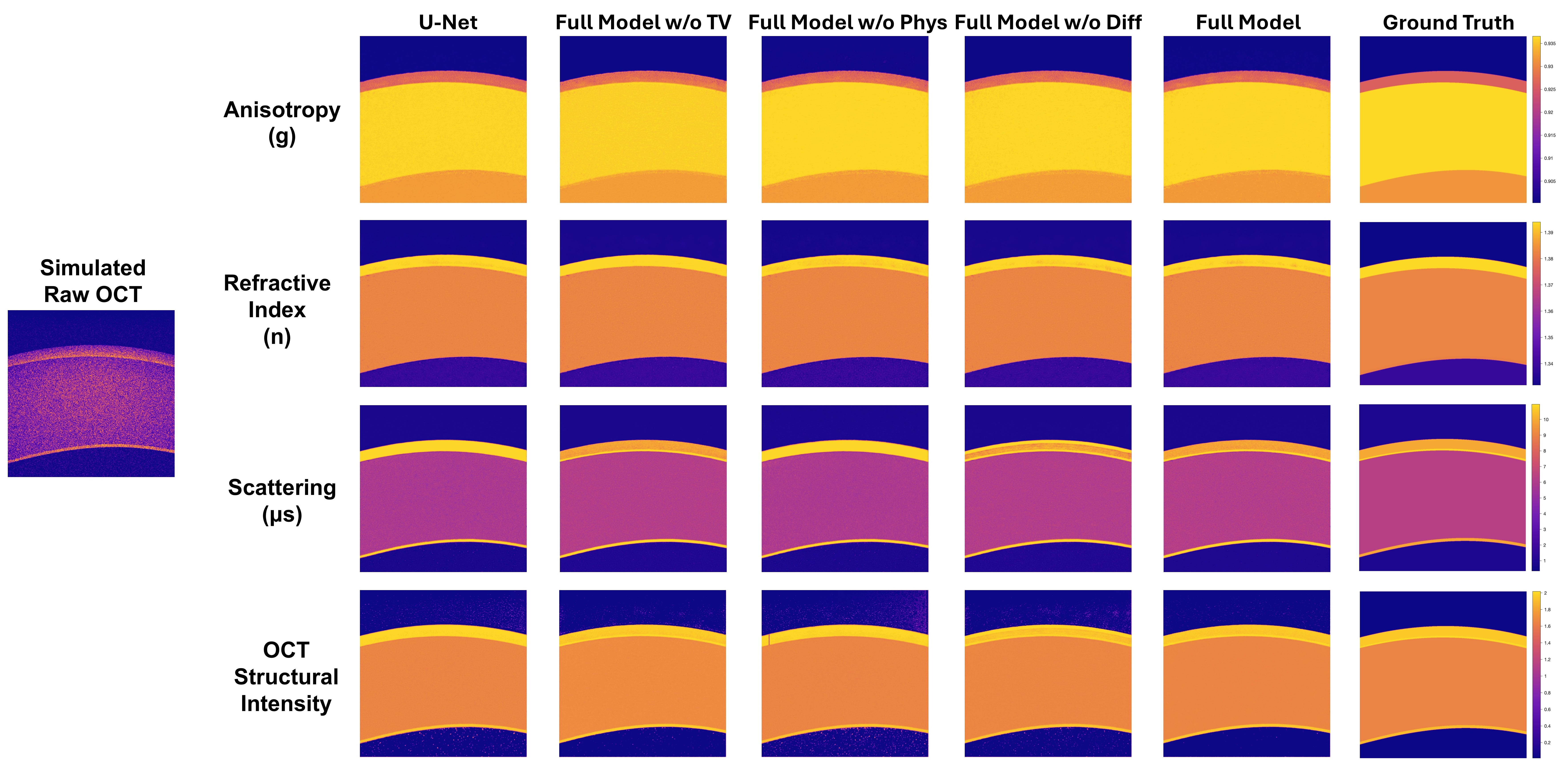}
    \caption{End-to-end reconstruction of OCT structural intensity and quantitative optical maps with ablation studies.}
    \label{fig:fig2}
\end{figure}

From the qualitative comparisons, removing diffusion regularization leads to incorrect $\mu_s$ outputs, as shown in figure~\ref{fig:fig2}. 
The recovered scattering profile no longer follows the five-layer distribution defined in the Monte Carlo modeling of light transport in multi-layered tissues (MCML) corneal model\cite{wang1995mcml}, and an additional artificial layer appears. 
The baseline U-Net underfits and reconstructs fewer than five layers, suggesting insufficient preservation of layer structure in the absence of appropriate priors.

All ablated variants---including the U-Net baseline and the full model without TV, diffusion, or physics-based constraints---exhibit speckle-like artifacts in the predicted OCT intensity $I_s$, particularly in the air and vitreous humor regions. 
In contrast, the proposed framework produces the cleanest and most clearly stratified layer boundaries.

\subsection{Quantitative Comparisons}
%\FloatBarrier
Table~\ref{tab:sr_quant_oct_us} and~\ref{tab:sr_quant_n_g} show superior quantitative performance of the proposed model across all metrics related to speckle-reduced OCT structural intensity $\hat{I}_{\mathrm{OCT}} $ and anisotropy $g$. 
These results indicate that multi-branch U-Nets combined with forward-model constraints improve reconstruction stability, noise suppression, and structural fidelity. Both PSNR and SSIM decrease significantly when diffusion regularization is removed, indicating that the absence of a statistical prior reduces robustness to noise. 
Removing the physics-based forward constraint further reduces reconstruction accuracy, with the structural intensity error increasing from $2.75 \times 10^{-3}$ to $1.83 \times 10^{-2}$. 
Therefore, pixel-level supervision alone is insufficient to constrain parameter coupling and hinder convergence towards physically plausible solutions.

The ablation results also reveal substantially higher errors in $\mu_s$ and $g$ when constraints are removed, whereas predictions of the refractive index $n$ change only slightly. 
This behavior may be attributed to the small dynamic range of corneal refractive index (approximately 1.32--1.40\cite{gong2020parametric}), which allows reasonable estimations even in the absence of explicit priors.

Finally, the baseline U-Net yields the lowest PSNR and SSIM and the highest MSE, revealing the limitations of purely data-driven learning. 
Specifically, the baseline model cannot jointly account for attenuation effects, parameter coupling, and noise amplification, and is therefore insufficient for reliable recovery of tissue optical properties from raw OCT intensity alone.

\begin{table*}[t]
\centering
\caption{Quantitative comparison on OCT structural intensity and scattering coefficient $\mu_s$. \textbf{Bold}: best.}
\label{tab:sr_quant_oct_us}
\begin{tabular}{l c c c c c c}
\hline\hline
 & \multicolumn{3}{c}{OCT Structural Intensity} & \multicolumn{3}{c}{$\mu_s$} \\
\cline{2-7}
Model 
& PSNR~$\uparrow$ & SSIM~$\uparrow$ & MSE~$\downarrow$
& PSNR~$\uparrow$ & SSIM~$\uparrow$ & MSE~$\downarrow$ \\
\hline
Full Model 
& \textbf{31.65} & \textbf{0.94} & $\boldsymbol{2.75 \times 10^{-3}}$
& \textbf{28.03} & 0.70 & \textbf{0.23} \\
Full Model w/o Diffusion 
& 27.41 & 0.88 & $7.29 \times 10^{-3}$
& 26.46 & \textbf{0.73} & 0.33 \\
Full Model w/o Physics 
& 23.41 & 0.80 & $1.83 \times 10^{-2}$
& 24.09 & 0.70 & 0.56 \\
Full Model w/o TV 
& 28.74 & 0.92 & $5.3 \times 10^{-2}$
& 27.93 & 0.71 & \textbf{0.23} \\
Baseline U-Net 
& 25.04 & 0.85 & $1.2 \times 10^{-2}$
& 24.09 & 0.63 & 0.56 \\
\hline\hline
\end{tabular}
\end{table*}
\begin{table*}[t]
\centering
\caption{Quantitative comparison on refractive index $n$ and anisotropy $g$. \textbf{Bold}: best.}
\label{tab:sr_quant_n_g}
\begin{tabular}{l c c c c c c}
\hline\hline
 & \multicolumn{3}{c}{$n$} & \multicolumn{3}{c}{$g$} \\
\cline{2-7}
Model 
& PSNR~$\uparrow$ & SSIM~$\uparrow$ & MSE~$\downarrow$
& PSNR~$\uparrow$ & SSIM~$\uparrow$ & MSE~$\downarrow$ \\
\hline
Full Model 
& 28.76 & 0.66 & $6.54 \times 10^{-6}$
& \textbf{29.25} & \textbf{0.84} & $\boldsymbol{1.90 \times 10^{-6}}$ \\
Full Model w/o Diffusion 
& 28.79 & 0.66 & $6.49 \times 10^{-6}$
& 26.82 & 0.80 & $3.33 \times 10^{-6}$ \\
Full Model w/o Physics 
& 28.90 & 0.65 & $6.32 \times 10^{-6}$
& 28.57 & 0.81 & $2.23 \times 10^{-6}$ \\
Full Model w/o TV 
& 28.78 & 0.66 & $6.50 \times 10^{-6}$
& 24.31 & 0.75 & $5.93 \times 10^{-6}$  \\
Baseline U-Net 
& \textbf{29.29} & \textbf{0.67} & $\boldsymbol{5.78 \times 10^{-6}}$
& 27.89 & 0.78 & $2.60 \times 10^{-6}$ \\
\hline\hline
\end{tabular}
\end{table*}

%\FloatBarrier
\section{CONCLUSION}

In summary, we present a physics-regularized end-to-end framework for OCT inverse reconstruction that simultaneously recovers the tissue optical properties and structural information from speckle-reduced OCT intensity in a single forward pass. By embedding the physics-based forward model and diffusion-based priors, the proposed approach effectively addresses attenuation, speckle noise, and parameter coupling that commonly compromise reconstruction accuracy. Experimental results demonstrate reliable recovery of structural information directly from MCML simulated OCT Bscans of corneal tissues, highlighting the potential of physics-informed deep learning for quantitative, label-free tissue visualization.

\acknowledgments % equivalent to \section*{ACKNOWLEDGMENTS}       
 
This work was supported by National Institute of Health Grant Award No. 1R01EY032127 (PI: Jin U. Kang), the study was conducted at Johns Hopkins University.

% References
\bibliography{report} % bibliography data in report.bib
\bibliographystyle{spiebib} % makes bibtex use spiebib.bst

\end{document}